\documentclass{article}

\usepackage{microtype}
\usepackage{graphicx}
\usepackage{subfigure}
\usepackage{booktabs} 

\usepackage{hyperref}

\usepackage[accepted]{arxiv}

\usepackage{amsmath}
\usepackage{amssymb}
\usepackage{mathtools}
\usepackage{amsthm}

\usepackage[capitalize,noabbrev]{cleveref}

\theoremstyle{plain}

\theoremstyle{definition}

\theoremstyle{remark}

\usepackage[textsize=tiny]{todonotes}

\icmltitlerunning{Diffusion-based Planning with Learned Viability Filters}

\usepackage{amsmath,amsfonts,bm}

\newcommand{\rvmu}{{\mu}}
\newcommand{\rvtau}{{\tau}}
\newcommand{\rvsigma}{{\Sigma}}
\newcommand{\rveps}{{\eps}}

\def\eqref#1{equation~\ref{#1}}

\def\1{\bm{1}}

\def\eps{{\epsilon}}

\def\rvc{{c}}

\def\rvz{{z}}

\DeclareMathAlphabet{\mathsfit}{\encodingdefault}{\sfdefault}{m}{sl}
\SetMathAlphabet{\mathsfit}{bold}{\encodingdefault}{\sfdefault}{bx}{n}

\newcommand{\meanp}[2]{\mathbb{E}_{#1} \left\lbrack #2 \right\rbrack}

\newcommand{\norm}[1]{\left\lVert#1\right\rVert}

\newcommand{\VF}{\ensuremath{\mathit{VF}}}

\begin{document}

\twocolumn[
\icmltitle{Diffusion-based Planning with Learned Viability Filters}




\icmlsetsymbol{equal}{*}

\begin{icmlauthorlist}
\icmlauthor{Nicholas Ioannidis}{ubc}
\icmlauthor{Daniele Reda}{equal,ubc}
\icmlauthor{Setareh Cohan}{equal,ubc}
\icmlauthor{Michiel van de Panne}{ubc}
\end{icmlauthorlist}

\icmlaffiliation{ubc}{Department of Computer Science, University of British Columbia, Vancouver, BC, Canada}

\icmlcorrespondingauthor{Nicholas Ioannidis}{nickioan@cs.ubc.ca}


\vskip 0.3in
]



\printAffiliationsAndNotice{\icmlEqualContribution} 

\begin{abstract}

Diffusion models can be used as a motion planner by sampling from a distribution of possible futures.
However, the samples may not satisfy hard constraints that exist only
implicitly in the training data, e.g., avoiding falls or not colliding with a wall.
We propose learned {\em viability filters} that 
efficiently predict the future success of any given plan, i.e., diffusion sample, and thereby enforce an implicit future-success constraint. 
Multiple viability filters can also be composed together. 
We demonstrate the approach on detailed footstep planning for challenging 3D human locomotion tasks, showing the effectiveness of viability filters 
in performing online planning and control for box-climbing, step-over walls, and obstacle avoidance.
We further show that using viability filters is significantly faster than guidance-based diffusion prediction.

\end{abstract}

\section{Introduction}

\begin{figure*}[!t]
\centering
\includegraphics[width=0.8\textwidth]{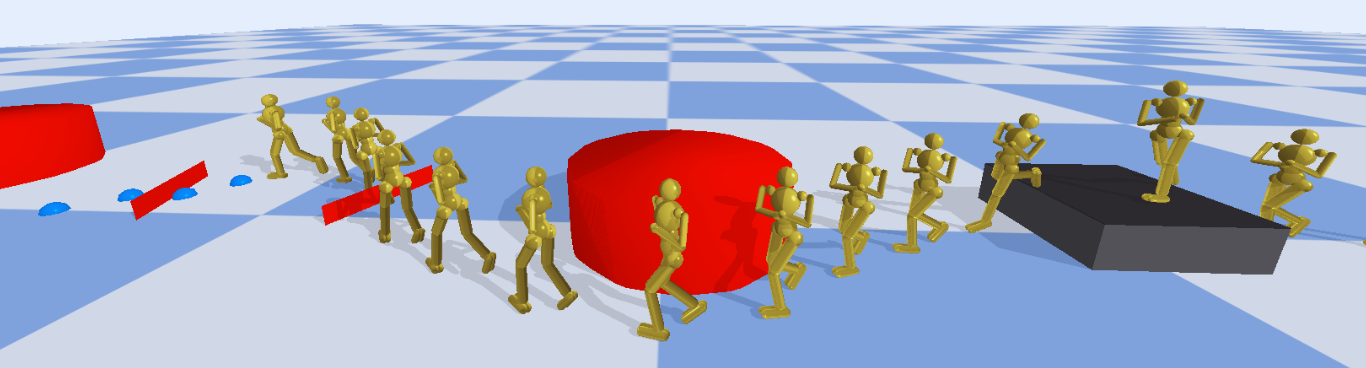}
\caption{The footstep-based planner with viability filters enables a humanoid to climb platforms, step-over hurdles, and avoid obstacles.}
\label{fig:demo}
\end{figure*}


Diffusion models offer a promising approach for planning and control.
Given suitable motion sequences for use as training data, and a current state,
a diffusion model can be used to model the distribution of plausible short-term futures (states and actions over some fixed horizon) from the current state.
This can then be used as an online controller using regular replanning, analogous to model-predictive control (MPC) methods.
Specifically, a sample from the diffusion model produces a plausible plan;
the first actions of the plan are then executed, resulting in a new state; and lastly, this iterative process repeats.


In practice, the approach just described comes with a number of limitations.
First, the diffusion model is unlikely to be able to perfectly model hard constraints that might exist implicitly
in the data, e.g., that of a foot always having sufficient clearance to traverse an obstacle.
Second, if the environment is stochastic or partially observable, then even the exact reproduction of an action sequence
that was successful in the past may provide no guarantees regarding its current success.
Third, distribution shifts can be problematic, e.g., the test-time environment may differ from the environment in which the training sequences were gathered.
Each of these issues can lead the diffusion model to propose infeasible plans, e.g., a humanoid taking a step that results in an eventual fall.


A possible solution is to filter infeasible plans, using a rejection test to discard such samples.
However, the issues described above generally preclude a fast-and-simple rejection test --- many feasibility constraints
are implicit in nature, and a stochastic or partially-observable environment means that it is difficult to predict
what will actually happen. Lastly, the immediate-future states and actions might be feasible and problem-free, but may nevertheless lead
to inevitable failure in the future, i.e., they may commit the system to entering what is a `dead end' situation.


We introduce a {\em learned viability filter} ($\VF$) that can be used as a fast-and-effective filter for proposed plans, testing plans 
for future feasibility. This can be formally modeled by testing states for membership in the viability kernel of an environment, defined as the subset of initial states in the environment such that there always exists at least one (controllable) path to an indefinitely ``viable'' or feasible future~\cite{viabilityTheoryBook}. 
The concept of viability has further strong connections to safety and robustness in reinforcement learning (RL), e.g.,~\citet{viabilityRobustRL}.
Our learned $\VF$ approximates the true viability kernel via a Q-function that can be learned using offline value iteration, and performs best
with the addition of online value iteration.


The $\VF$ approach has key attributes that are difficult to achieve directly with the diffusion model.
It can learn from both positive and negative outcomes, whereas the diffusion model is trained on motion sequences with positive oucomes. 
The $\VF$-as-Q-function produces expected viability, and thus takes into account the stochastic nature of both the policy and the environment, as well as taking into future consequences.  The $\VF$ can be trained independently of the diffusion model, which allows for conditioning on different information, if available, as well as online composition of multiple $\VF$s. We demonstrate these features in the context of controlling a physics-based 3D humanoid performing challenging locomotion tasks.


It can be useful to situate our work on an axis that spans offline-compute vs online-compute.
At one end, diffusion-based controllers and RL controllers focus on learning for generation, i.e., offline training is used to develop 
control policies that, at runtime, can directly produce suitable actions for a given state.  At the other end of the spectrum, it is possible to focus entirely on online compute. For example, at any given point in time, sampling-based model-predictive control methods may generate a fixed number of fixed-horizon random-action plans that are then evaluated online (using model-based simulation) in order to  select the best available plan. Our learned $\VF$ approach sits in-between these two extremes and aims to leverage the benefits of both. 
A $\VF$-enabled diffusion-based controller allows for efficient generation of a diverse set of plausible plans (unlike MPC methods) while also not requiring the diffusion model to `completely get it right' (unlike pure diffusion-based controllers). 
$\VF$s further offer useful flexibility -- they can be composed together at inference time, and can be conditioned on task information or observations that are not availble to the diffusion model.  We shall further show that $\VF$s offer a significant speed advantage over guidance-based diffusion methods.

\section{Related Work}
\label{sec:related work}
\paragraph{Diffusion Models}
Diffusion models \citep{sohl2015deep, ho2020denoising, song2020denoising, nichol2021improved} are a class of generative models that approximate data distributions through an iterative denoising process and have shown remarkable success in generating images \citep{dhariwal2021diffusion, ramesh2022hierarchical, rombach2022high}, videos \citep{ho2022video, harvey2022flexible}, and audio \citep{kong2020diffwave}. Diffusion models allow for several conditioning approaches for conditional generation. Classifier-free guidance~\cite{ho2022classifier} explicitly trains a conditional diffusion model. Classifier guidance~\cite{dhariwal2021diffusion} uses gradients of a trained classifier to encourage samples of an unconditionally trained diffusion model to satisfy the condition. Several methods have been proposed for replacing observations in diffusion outputs~\cite{lugmayr2022repaint, ho2022video} and editing diffusion outputs with desired criteria ~\cite{meng2021sdedit, parmar2023zero}. 

Constraint-aware diffusion models extend conditional generation by introducing strict requirements necessary to be satisfied. The simplest approach, rejection sampling, filters out invalid samples but can be computationally expensive due to costly diffusion inference. It also becomes impractical in the absence of a world model or in noisy environments. Alternatively, samples can be guided to stay within the predefined boundaries using classifier guidance~\citep{pmlr-v235-naderiparizi24a} and iterative projection ~\citep{christopher2024constrained}.
Another line of work focuses on modifying the sampling to keep generated data within the constrained boundary \citep{lou2023reflected, fishman2024metropolis, liu2024mirror}. Other approaches employ diffusion bridges, stochastic processes designed to terminate within a specific constraint set, to solve this problem \citep{liu2023learning}.

\paragraph{Diffusion-based Planning}
Diffusion models have recently been applied to offline-RL planning. Diffuser~\citep{janner2022diffuser} treats planning for offline RL as a sequence modeling problem and trains a diffusion model to generate state-action trajectories, which are guided towards high-reward regions using inference-time guidance. 
Decision Diffuser~\citep{ajay2022conditional} trains a diffusion model to generate sequences of states and executes actions derived from a trained inverse dynamics model. It also uses classifier-free guidance for goal-conditioned generation.
Diffusion Policy (DP)~\citep{chi2023diffusion} generates sequences of actions conditioned on a small history of states, leading to improved inference-speed. Diffusion-QL~\citep{wang2022diffusion} enhances DP by incorporating a state-value term into the diffusion loss, using a jointly trained Q-function to guide sampling toward higher rewards. LDCQ~\citep{venkatraman2023reasoning} proposes a latent diffusion planner that trains a Q-function for filtering generated actions at inference time. DIAR~\citep{park2024diar} extends this by introducing a value function for improving the Q-learning, using both Q and value functions at inference to refine action selection.

Going beyond benchmark offline-RL tasks, DiffuseLoco~\citep{huang2024diffuseloco} and PDP~\citep{truong2024pdp} apply DP to robot and animation control respectively. Trace and Pace~\citep{rempeluo2023tracepace} uses a diffusion planner for guided pedestrian motion planning. A physics-based humanoid controller is then used for following the generated plans, and reconstruction guidance is used for added controlability. CTG~\cite{zhong2023guided} conditions diffusion-based vehicle planning on agent states and the map, using classifier guidance for road rule compliance. DJINN~\cite{niedoba2024diffusion} jointly models agent trajectories and applies classifier guidance based on scene semantics. Gen-Drive~\citep{huang2024gen} uses diffusion to model possible future scenarios and utilizes a trained reward model to facilitate decision making.

\section{Background}
\label{sec:background}
We model planning as a discounted Markov Decision Process (MDP) defined by the tuple $\langle \rho_0, \mathcal{S}, \mathcal{A}, \mathcal{T}, R, \gamma \rangle$, where $\rho_0$ is the initial state distribution, $\mathcal{S}$ and $\mathcal{A}$ are state and action spaces, $\mathcal{T}: \mathcal{S} \times \mathcal{A} \rightarrow \mathcal{S}$ is the transition function, $\mathcal{R}: \mathcal{S} \times \mathcal{A}  \times \mathcal{S} \rightarrow \ \mathbb{R}$ is the transition reward and $\gamma \in [0, 1)$ is the discount factor \citep{puterman2014markov}. The goal is to generate trajectories $\rvtau = (s_0, a_0, s_1, a_1, ..., s_{T}, a_{T})$ over a planning horizon $T$ such that the return $\mathcal{J}(\rvtau) := \sum_{t=0}^{T} \gamma^t r(s_t, a_t)$ is maximized.

Given a dataset of trajectories $\rvtau^0 \sim q(\rvtau^0)$, Diffusion probabilistic models (DDPMs) \citep{sohl2015deep, ho2020denoising} define a forward diffusion process $q(\rvtau^{i} | \rvtau^{i-1})$ that iteratively corrupts data through the addition of small amounts of Gaussian noise:
    \begin{align*}
         q(\rvtau^{i} | \rvtau^{i-1}) = \mathcal{N}(\rvtau^{i}; \sqrt{1-\beta^i}\rvtau^{i-1}, \beta^i{I})
    \end{align*}
where $i$ is the diffusion step and $\beta^{1,...,N}$ is a variance schedule indicating the amount of noise. When the forward transitions are small, the reverse process can be parameterized as a sequence of Gaussian distributions with fixed, time-dependent covariances $p_\theta(\rvtau^{i-1} | \rvtau^{i})$ that learn to remove noise from $\rvtau^i$ starting from pure Gaussian noise $\rvtau^N$:
    \begin{align*}
        p_\theta(\rvtau^{i-1} | \rvtau^{i}, \rvc) = \mathcal{N}(\rvtau^{i-1}; \rvmu_\theta(\rvtau^i, i, \rvc), \rvsigma^i)
    \end{align*}
where $\rvc$ is some conditioning signal or auxiliary input and the covariance is set as $\rvsigma^i = {\sigma^i}^2 \mathbf{I}$ similar to DDPMs. Following Diffuser~\cite{janner2022diffuser}, we use the noise-estimation parameterization for the diffusion planner and directly approximate the noise estimate $\rveps$ instead of the mean-estimate $\rvmu$ by optimizing the following objective:
    \begin{align}
        \mathcal{L} = \meanp{(\rvtau^0, \rvc) \sim q(\rvtau^0, \rvc), i \sim [1, N], \rveps \sim \mathcal{N}(\mathbf{0}, \mathbf{I})}{
        \norm{
            \rveps - \rveps_\theta(\rvtau^i, i, \rvc)
        }^2
        }.
        \label{eq:diffusion_loss}
    \end{align}
With the trained reverse process noise function approximator $\rveps_\theta$, denoising at every step is done by:
    \begin{align*}
        \rvtau^{i-1} = \frac{1}{\sqrt{\alpha^i}} \left ( \rvtau^i - \frac{1-\alpha^i}{\sqrt{1-\bar{\alpha}^i}} \rveps_\theta(\rvtau^i, i, \rvc) \right ) + \sigma^i \rvz
    \end{align*}
where $\alpha^i = 1 - \beta^i$, $\bar{\alpha}^i = \prod_{k=1}^{i} \alpha^i$ and $\rvz \sim \mathcal{N}(\mathbf{0}, \mathbf{I})$.

\emph{Classifier guidance} allows for conditioning a pre-trained diffusion model by pushing the samples towards desired conditions at inference time. In planning, we can guide samples towards areas that increase the return for any desired task at inference time. With $\mathcal{J}_\phi(\rvtau^i)$ being the return over noisy trajectory $\rvtau^i$, the denoising operation is modified at $i$ simply by adding $\nabla_\phi\mathcal{J}(\rvtau^i)$ to the mean estimate derived from the output of the diffusion model $\rveps_\theta$. For details on diffusion guidance, refer to~\ref{sec:appendix-guidance}.

\begin{figure}[H]
  \centering
  \includegraphics[width=0.38\textwidth]{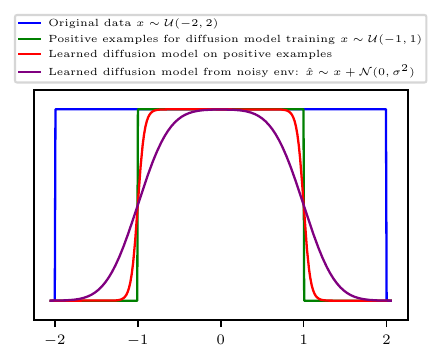}
  \caption{Toy example with unnormalized probability density functions. 
  Learned diffusion models do not fully respect hard constraints, in particular when trained on noisy data,
  e.g., observations from a stochastic environment. }
  \label{fig:toy_env}
\end{figure}

\paragraph{Limitations of Diffusion Planners}
We use Fig.~\ref{fig:toy_env} to highlight some of the limitations of diffusion models. Consider a 1D environment with an implicit constraint $C = [-1,1]$. If a drawn sample $x$ lies in $C$ then this outcome would be a success, otherwise a failure. The goal is to train a diffusion model that generates samples within this implicit constraint $C$. In order to capture the hidden constraints samples are drawn form a wider uniform distribution $U(-2,2)$ and the successful outcomes (which result to $U(-1,1)$) are used to train the diffusion model. 

\begin{figure*}[!ht]
  \centering
  \includegraphics[width=0.85\textwidth]{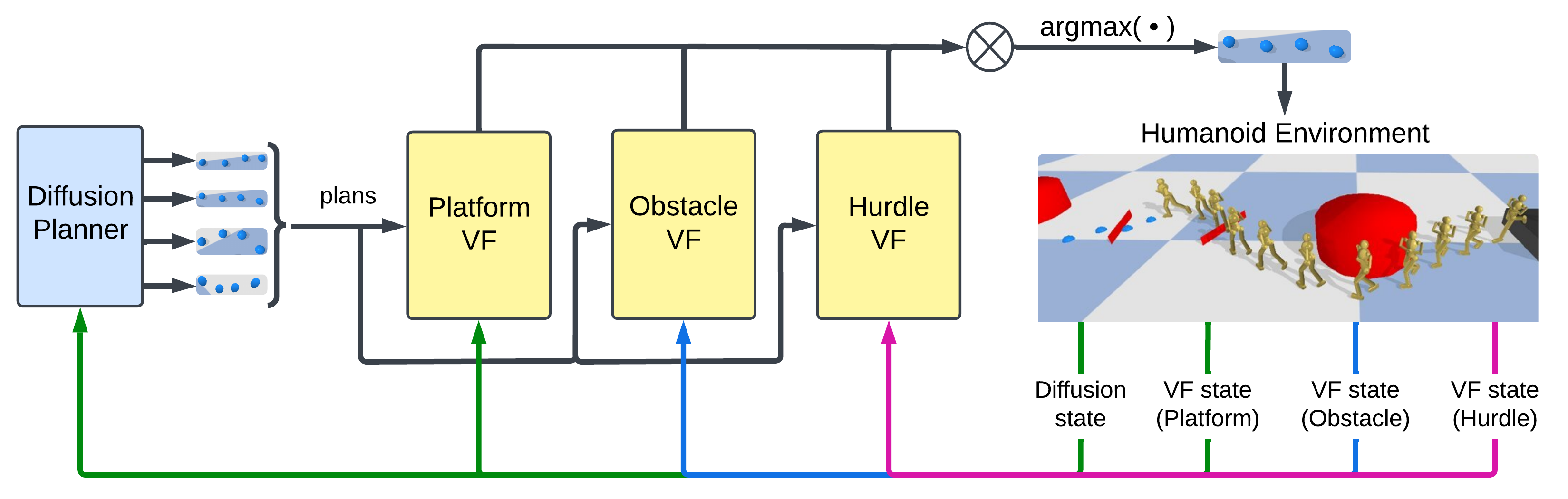}
  \caption{\textbf{Inference pipeline of our compositional environment.} A diffusion planner generates a set of possible plans given the diffusion state (including the character state, the height map and the waypoint). Given task-relevant information as input to each $\VF$, each plan is evaluated by all $\VF$s and a viability value is associated to each plan according to Eq.~\ref{eq:composition_vf}. The plan with the highest viability is executed in the environment by the controller. Note that the inference pipeline is similar in the case of a single task, as the diffusion planner is the same for all, but the generated plans are evaluated on a single $\VF$.}
  \label{fig:inference}
\end{figure*}

The generated samples from the diffusion model then mostly capture $C$, but some samples still end-up \textit{leaking} outside the boundaries. This already assumes an ideal scenario, with clean data and a deterministic environment. In many cases environments are stochastic in nature and we may further be restricted to work with partial or noisy observations. In our example we can consider a stochastic environment with perturbations or observational noise sampled from $N(0,\sigma^2)$, which is added to the sample value such that $\hat{x}\sim x+ N(0,\sigma^2)$ before evaluating its outcome. With respect to the given constraint, this results in significant  misclassifications with respect to the constraint and make the resulting data distribution (and the diffusion model that is trained on it) do poorly at maintaining the constraint, $C$. In sequential decision making, we are further concerned with future failure, which is not captured in our toy example.

\section{Method}

Our planning framework consists of
a diffusion-based planner that is sampled to propose batches of short-term plans, and 
one-or-more viability filters (VF) that evaluate the feasibility of any given plan. 
The most viable selected plan is then passed to the environment. 
At runtime, in a receding-horizon predictive-control fashion, we replan after the first action.
The specific details of our test environment will be described shortly (\S\ref{sec:env}).

Our diffusion-based planner builds on the basic design of Diffuser~\citep{janner2022diffuser} optimizing the following loss:
    \begin{align*}
        \mathcal{L} = \meanp{(p^0, \rvc) \sim q(p^0, \rvc), i \sim [1, N], \rveps \sim \mathcal{N}(\mathbf{0}, \mathbf{I})}{
        \norm{
            \rveps - \rveps_\theta(p^i, i, \rvc)
        }^2
        }
    \end{align*}
where $p^i$ is the noisy plan at diffusion step $i$, and $\rvc$ is the conditioning signal.

The Viability Filter~($\VF$) is approximated as a Q-function and estimates the discounted cumulative reward given the current state $s_t \in S$ and a plan $p_t \in P$ such that:
\begin{equation}
    \VF(s_t,p_t) = \mathbb{E}[r_{t+1} + \gamma r_{t+2} + \cdots | s_t,p_t].
\end{equation}
Here, $r_t$ is a binary reward that functions equivalent to an `alive' reward applied at any environment step that successfully does not fail,
according to a given failure criterion.
$\VF(s_t,p_t)$ has a maximum value $Q_{\max}=1/(1-\gamma)$, corresponding to an indefinitely viable plan. 
The value of $\VF$ can be used to define a diffusion-based control policy in several ways, the simplest being to generate plans until a plan is found that meets a given threshold, e.g., $\VF(s_t,p) \geq \beta Q_{\max}$, where $\beta$ defines a threshold to be exceeded, e.g., $\beta=0.95$.
In practice, we use a policy described by $p_t = \arg\max_{p_i} Q(s_t,p_i), i\in [1,N]$, where $N$ samples are drawn from the diffusion model, i.e., $N$ plans are generated and tested.

\section{Environment \label{sec:env}}

We test our planning framework on challenging humanoid locomotion scenarios, including step-up/down and step-over situations that need to be precisely executed to avoid falling or tripping. 
We plan at the level of target footsteps, which serve as the action space of the overall humanoid environment. 
The low-level details of target footstep execution are then allocated to a pre-trained RL-based locomotion
controller. For this, we use a control policy trained according to the details of ALLSTEPS~\cite{2020-SCA-ALLSTEPS}, which produces joint torques over time and is conditioned on the next two footstep locations. 
For the purposes of our work here, the behavior of the low-level controller is treated as a black-box, i.e.,  part of the environment.  Both the diffusion model and the $\VF$ work only with target footstep sequences, although they do see the full body state, as that is critical to making suitable footstep planning decisions.

We define a \textit{plan} $p = (f_0, f_1,...,f_{n_p - 1})$ as a set of $n_p$ consecutive footstep locations $f_t$. Each footstep is denoted as $f_t = \langle f_{t_x},f_{t_y},f_{t_z} \rangle$ corresponding to the relative distance of the $t_{th}$ target footstep location in the plan with respect to the humanoid character. For this work we choose $n_p=4$ which corresponds to 4 consecutive target footstep locations.
We further define the notion of a \textit{trajectory} --- this refers to a fixed and complete sequence of footstep locations (i.e., extending beyond the time horizon of the episode) that is generated prior to the start of the episode. 
Trajectories will be used to collect training data on long fixed sequences, while plans provide a short-term planning horizon, consisting of four footstep locations, and they form the action space of our diffusion-space planner-plus-$\VF$.

We design three tasks, each aimed at addressing different planning challenges:
a \textit{step-up platform} task,
a \textit{step-over hurdle} task, and
a \textit{obstacle avoidance} task.
Specifics of the three tasks are described below.

\paragraph{Step-up platform task}
The \textit{step-up platform} task consists of {$2\times2$}~m cubes with height in the {$[10, 65]$}~cm range. This range was determined based on the height distribution used to train the controller, which observed a maximum height of 60~cm. We increased the upper limit to 65~cm to examine the resulted behavior when the task exceeds the capabilities of the controller. To place the platforms, we first compute a complete trajectory using our procedural generator, then we model the resulting trajectory curve as a B-spline, and finally randomly position the platforms along this curve.

\paragraph{Step-over hurdle task}
The goal of this task is to evaluate the capacity of the viability filter to adapt to situations where the planner is unaware of the constraints (i.e., the hurdles are not visible to the planner). To this end, we place hurdles of zero length that are not visible in the environment state of the planner. The hurdles have a height ranging from 25~cm to 35~cm.
Although the height of the hurdle was chosen to fall within the foot clearance range observed while walking on flat terrain, it still poses a risk of tripping or falling without careful planning. Similarly to the platform environment, the obstacles are randomly placed along a procedurally generated trajectory and oriented perpendicular to the curve.

\paragraph{Obstacle avoidance task}
Similar to the \textit{step-up hurdle} task, the \textit{obstacle avoidance} task requires the character to avoid obstacles not visible to the planner.
Obstacles are represented as cylindrical objects with a radius ranging from 0.5~m to 1.5~m. In order to traverse the terrain successfully, the planner together with the viability filter need to create a trajectory that maneuvers around the cylindrical obstacles.

\paragraph{Procedurally Generated Trajectories}
 Data collection begins with procedural trajectory generation, which provides diverse footstep trajectories. Starting with a flat terrain and an initial heading angle $\phi_{t}$, which describes the direction at which the character will move, each consecutive target footstep along the trajectory is generated by first sampling a new radial footstep distance $\Delta r\sim U(\Delta r_{min},\Delta r_{max})$ and a heading angle offset $\Delta\phi\sim U(-\Delta\phi_{max},\Delta\phi_{max})$. 
To encourage a natural walking gait, we rotate the target step's coordinates depending on whether the left or right leg will be used to reach it.

\section{Experimental Setup}

\subsection{Diffusion Planner}

\begin{figure}[h]
  \centering  \includegraphics[width=0.35\textwidth]{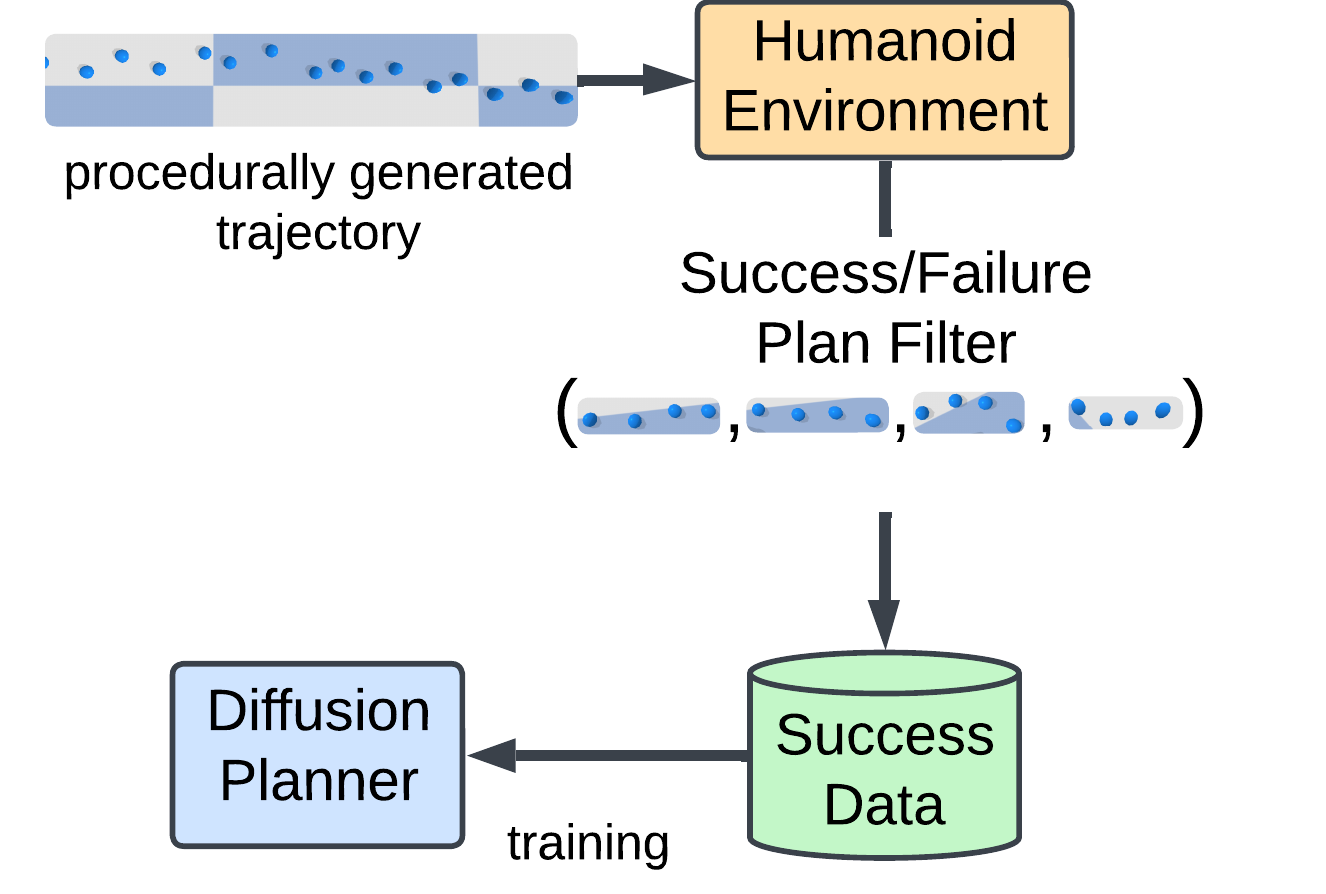}
  \caption{Training procedure. A humanoid controller follows procedurally-generated footstep trajectories. Each 4-footstep window is categorized as a success or failure. Successful plans are used to train the Diffusion Planner.}
\label{fig:diffusion_train}
\end{figure}

We train the diffusion planner on procedurally generated trajectories in both flat terrain and the step-up platform task as shown in Fig.~\ref{fig:diffusion_train}. Each trajectory effectively contains multiple plausible plans of four-footstep sequences. We condition the diffusion model on the egocentric local heightfield, the character state and the user-defined waypoint goal. The heightfield is a $32\times32$ matrix that represents the height of a {$5\times6$}~meters area in the environment, shifted forward to represent 4.5~m in front of the character and 0.5~m behind it, centered along its width, and oriented towards the next user-defined target waypoint. The character state includes orientation of the character relative to the global coordinate frame, joint angles and angular velocities, body velocities, binary foot contacts, and relative position for each foot. Finally, the waypoint corresponds to a relative 3D goal point in space that the character is tasked to reach.

In the platform task, waypoints align with platform centers, while in flat terrain, they are randomly placed. This setup enables learning from diverse trajectories, with and without platforms. A key challenge is ensuring the character interacts with and traverses platforms, especially when heights exceed the controller's training distribution. To address this, we specifically place the platforms along the procedurally generated path, ensuring the character consistently attempts to traverse the platforms rather than bypassing them. A plan is successful if the character avoids early termination (i.e., falling) before completion. If the character falls at step $f_t$, any plan including $f_t$ is marked as failed, while earlier plans remain successful. Falling is detected when the height difference between any foot and the torso drops below a threshold. Details on the diffusion planner architecture and training is available in our supplementary materials.

\subsection{Viability Filter}

\paragraph{State Representation} While the character state is included in the viability filter state across all tasks, additional components are task-specific. This allows us to reduce the size of the network by excluding irrelevant information that does not contribute to the task.
In the step-up platform task, the state also includes the same heightfield given to the planner;
in the step-over hurdle task, we include the hurdle's location and angle;
lastly, in the obstacle avoidance task, we provide the location and radius of the cylindrical object zone, as well as the waypoint location. The waypoint location is included in the viability filter state only in the obstacle avoidance task, as the character is required to deviate from the waypoint to avoid the obstacle.

\begin{figure}[!h]
  \centering
\includegraphics[width=.49\textwidth]{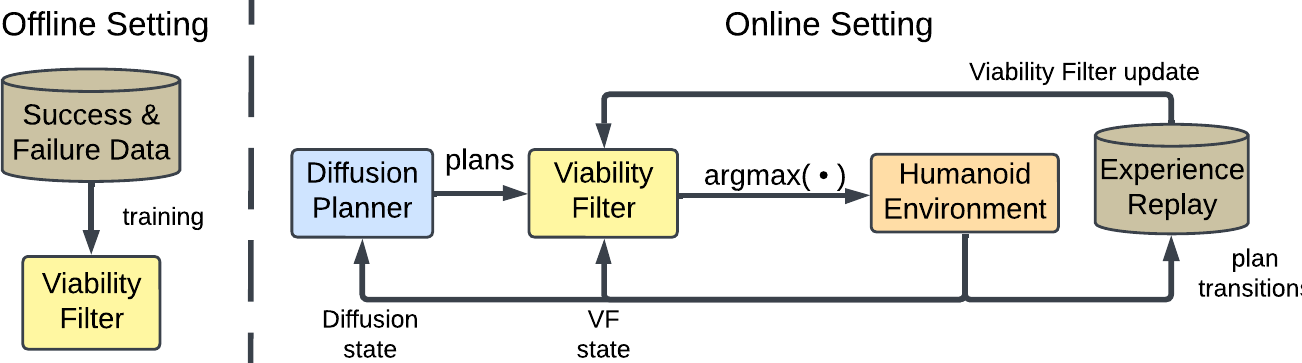}
  \caption{Training procedure for the Viability filter in the Offline (left) and Online (right) settings. 
  In the Online case the best proposed plan from the Diffusion Planner is evaluated and stored. Transitions in the Experience Replay is used in an Online-RL fashion to update the Viability Filter}
  \label{fig:vf_training}
\end{figure}

\paragraph{Training \& Architecture}
We propose two methods for training the $\VF$: (1) using offline data generated via the procedural trajectory generator, and (2) online with a frozen diffusion model generating the plans instead. Fig.~\ref{fig:vf_training} shows an overview of the training of the $\VF$.

In the offline case, when collecting data from the procedural generator, we compute the returns for each plan $p_t$ such that $R_{p_t} = \sum_{k=t}^{t+N} \gamma^{k-t} r_{k}$, where $\gamma = 0.75$. The objective function in the offline case minimizes the difference between the observed and estimated return:
\begin{equation}
    L_{\VF} = \frac{1}{2}(R_{s_t,p_t} - \VF_{\theta}(s_t,p_t))^2.
\end{equation}

In the online case instead, similarly to DQN~\cite{mnih2013playing}, transitions are stored in an experience replay and gradients are computed through a batch of samples using the Bellman backup function:
\begin{equation}
\begin{aligned}
    y_i &= r_i + \gamma \max_{p} \hat{\VF}(s_{i+1}, p_{i+1}), \\
    L_{\VF} &= \left(y_i - \VF(s_{i}, p_{i})\right)^2
\end{aligned}
\label{eq:bellman}
\end{equation}
where $\hat{\VF}$ is a target network introduced for stability purposes.
The main challenge with Eq.\ref{eq:bellman} is estimating the plan that maximizes the return (i.e. $\max_{p} \VF(s, p)$) since the plan space $P$ is neither discrete nor we can sample some expected plan as it is usually seen in RL settings.
Instead, we sample $N$ plans from the diffusion planner such that $\tilde{P}_{i+1}=(p_{i+1}^1, p_{i+1}^2, \cdots, p_{i+1}^N)$ represents a set of plans. Given a large enough value of $N$, $\tilde{P}_{i+1} \simeq P$ and as a result Eq.~\ref{eq:bellman} becomes:
\begin{equation}
\begin{aligned}
    y_i &= r_i + \gamma \max_{p_{i+1}\in\tilde{P}_{i+1}} \hat{\VF}(s_{i+1}, p_{i+1}) \\
    L_{\VF} &= \left(y_i - \VF(s_{i}, p_{i})\right)^2.
\end{aligned}
\label{eq:bellman_updated}
\end{equation}
A breakdown of the online training process and architecture for the viability filters is available in the supplementary material.

\section{Results}

We evaluate the diffusion planner and the $\VF$ on each of the three individual tasks involving footstep planning, as well as a composition of the tasks.

\subsection{Platform Task}
We first evaluate our framework on the platform task by measuring the likelihood of the character successfully traversing a platform (climbing up and down) without falling. The evaluation is conducted on platform heights ranging from 10~cm to 65~cm. We report results from three settings: (1) using only the diffusion model, (2) training the viability filter offline with data from the procedural generator, and (3) training the viability filter online using the pre-trained diffusion model.
We also report the likelihood of successfully traversing a platform by following the procedurally generated trajectory as an indication of the difficulty to succeed on the task without planning. Results are presented in Fig.~\ref{fig:box success rates}. Tabular values are in the supplementary material in Tab~\ref{tab:merged_success_rates}.

As presented in Fig.~\ref{fig:box success rates}, using the diffusion planner on its own improves performance across all heights compared to the procedural generator, as expected. Performance decreases as the height of the platform increases, and the task becomes harder.
When applying the offline-trained $\VF$, we observe an additional increase in performance, although for tall platforms (65~cm), the character fails most of the time. Adding the online-trained $\VF$ provides a significant improvement of performances on the hardest height settings, and additional overall consistency across the full range of platform heights. The advantages of the online $\VF$ over the offline $\VF$ are twofold: first, the data in the offline setting comes from the procedural generator whereas in the online setting it comes directly from the trained diffusion model; and, second, in the online setting the $\VF$ is trained with a temporal-difference error, which allows the full distribution of future plans to be taken into account. This is relevant because the true viability of a plan during inference does change if replanning occurs, something that is not captured in the offline setting, being trained on MSE.

\begin{figure}[htbp]
    \centering
    \includegraphics[width=1.\linewidth]{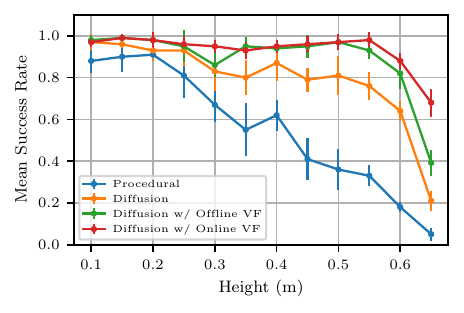}
    \caption{Mean success rate and standard deviation per platform for 10~cm to 65~cm heights using the procedurally generated trajectories, the diffusion planner only, and the diffusion planner with the offline and online trained Viability Filter, averaged across 20 episodes for 5 different trials.}
    \label{fig:box success rates}
\end{figure}

\subsection{Hurdle and Obstacle Tasks}
It is well known that training diffusion models may require a significant volume of relevant data which is not always feasible to acquire, and they struggle to generalize without this. We demonstrate that our approach allows the diffusion planner to generalize to new downstream tasks via an updated $\VF$ alone, which selects successful plans.  The $\VF$ can then also be conditioned on different observations. We exploit that here, where the $\VF$ is conditioned with relevant information regarding the hurdle and obstacle locations, whereas the diffusion model is oblivious to this.

We train a $\VF$ on the Hurdle task both on the online and the offline setting. For the offline setting we collect data using the procedural trajectory generator. During evaluation, waypoints are positioned at the center of each obstacle to ensure that the diffusion planner attempts to traverse them. We report our results in Tab.~\ref{tab:success_rates_hurdle}. We also include the success rate of naively using only the diffusion planner, which is not aware of the hurdles location, to demonstrate the difficulty of the task. Introducing the $\VF$ makes it possible for the character to traverse these terrains with higher success rate. Notably, the additional performance boost observed with the online method can be attributed to learning from the diffusion planner's observed distribution and adapting the $\VF$'s estimates using online data.

\begin{table}[tbhp]
\caption{Mean success rate and standard deviation per hurdle height using the diffusion planner only, and the diffusion planner with the offline and online trained Viability Filter, averaged across 20 episodes for 5 different trials.}
\resizebox{\columnwidth}{!}{%
\begin{tabular}{c|c c c}
\toprule
\textbf{Hurdle} & \textbf{Diffusion} & \textbf{Diffusion} & \textbf{Diffusion} \\
\textbf{Heights (cm)} & & \textbf{VF-Offline} & \textbf{VF-Online}\\
\hline
25 & 0.52 $\pm$ 0.13 & 0.93 $\pm$ 0.08 & 1.00 $\pm$ 0.00 \\
30 & 0.26 $\pm$ 0.12 & 0.85 $\pm$ 0.05 & 0.96 $\pm$ 0.04  \\
35 & 0.13 $\pm$ 0.12 & 0.70 $\pm$ 0.11 & 0.83 $\pm$ 0.07 \\
\bottomrule
\end{tabular}
}
\label{tab:success_rates_hurdle}
\end{table}

Unlike the previous tasks, where the character needs to traverse either the platform or hurdle, the obstacle task, instead, requires the character to avoid and maneuver around specified obstacles. This makes collecting offline data challenging, as placing zones along a procedurally generated path often guarantees failure. Because of this, we train the $\VF$ solely in the online setting.

Obstacle avoidance is a well-explored problem in motion-planning literature. Regarding diffusion models, a common strategy is to use classifier-based guidance. To compare our method with such approaches, we follow the work of~\citet{karunratanakul2023guided}, which uses the Signed Distance Function (SDF) as an obstacle avoidance guidance term. We adapt this guidance term for our setting as follows:
\begin{equation}
    \mathcal{J}(p) = \sum_i -\text{clipmax}(\text{dist}(f_t, o), \hat{r}_o) \quad \text{for } f_t \in p,
\end{equation}
where $\text{dist}$ refers to the Euclidean distance between $f_t$ and the center of the next obstacle $o$, and $c$ is the safe distance from the obstacle. We define $\hat{r}_o$ as $\hat{r}_o = r_o + \epsilon$, where $r_o$ is the radius of the obstacle, and $\epsilon$ is an empirically determined tolerance.

We report our evaluations in Tab.~\ref{tab:success_rate_obstacle}, including the success rate of the blind diffusion model, which represents the likelihood of  achieving obstacle avoidance by chance alone while reaching to the target waypoint. Our method significantly outperforms the guidance-based method, as the $\VF$ not only models plans that steer the character away from the obstacle but also evaluates plans based on their feasibility within the controller's capabilities.

\begin{table}[bth]
\caption{Mean success rate and standard deviation per obstacle radius size using the diffusion planner only, the diffusion planner with classifier-based guidance and the diffusion planner with the Viability Filter, averaged across 20 episodes for 5 different trials.}
\resizebox{\columnwidth}{!}{%
\begin{tabular}{c|c c c}
\toprule
\textbf{Obstacle} & \textbf{Diffusion} & \textbf{Diffusion} & \textbf{Diffusion} \\
\textbf{Radius (m)} & & \textbf{Guidance} & \textbf{VF-Online}\\
\hline
0.5 & 0.19 $\pm$ 0.02 & 0.67 $\pm$ 0.11 & 0.99 $\pm$ 0.02\\
1.0 & 0.11 $\pm$ 0.06 & 0.64 $\pm$ 0.05 & 0.99 $\pm$ 0.02 \\
1.5 & 0.00 $\pm$ 0.00 & 0.63 $\pm$ 0.04 & 0.99 $\pm$ 0.02 \\
\bottomrule
\end{tabular}
}
\label{tab:success_rate_obstacle}
\end{table}

\subsection{Effect of Sample Size \& Inference Speed}
\begin{figure}[htbp]
    \centering
    \includegraphics[width=1.\linewidth]{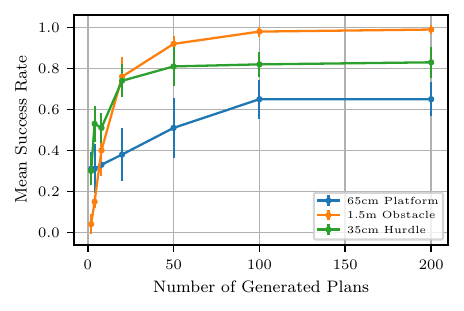}
    \caption{Mean success rate and standard deviation across the number of diffusion generated plans with the 
    Online Viability Filter during inference for the 65~cm platform task, 1.5~m obstacle and 35~cm hurdle, 
    averaged across 20 episodes and 5 different trials.}
    \label{fig:samples_vs_performance}
\end{figure}

An important parameter in our system is the number of samples generated by the diffusion planner since it directly affects the diversity of the potential plans to be evaluated by the $\VF$.
In Fig.~\ref{fig:samples_vs_performance} we evaluate how the number of generated samples during inference affects performance of the online $\VF$ across the most challenging setting for each task.
We observe that when the number of plans generated is small, the success rate of using the $\VF$ matches closely to that of only using the diffusion planner. This is expected, as a low number of plans provides insufficient choice to the $\VF$ to find the best plan to select. As the number of generated plans increase to 50, performance across all tasks reaches a plateau. 

We also report the inference speed of a single plan generation and evaluation across different number of samples and compare to that of a single guided sample. Values are provided in Tab.~\ref{tab:inference_speed}. Our method provides a significant inference-time speed-up as compared to classifier-based guidance. We further note that classifier-based guidance can be particularly challenging to tune.

\begin{table}[!htbp]
\caption{Inference runtime cost for a generation-evaluation step with the Diffusion Planner and Viability Filter for a batch of 100,200, and 400 plans compared to a single guidance step averaged across 20 iterations.}
\resizebox{\columnwidth}{!}{%
\begin{tabular}{c|c c c c}
\toprule
\textbf{Method} & \textbf{Guidance} & \textbf{VF} & \textbf{VF} & \textbf{VF}  \\
& \textbf{(1 sample)} & \textbf{(100 samples)} & \textbf{(200 samples)} & \textbf{(400 samples)}  \\
\hline
Time (s) & 0.387 & 0.105 & 0.111 & 0.123 \\
\bottomrule
\end{tabular}
}
\label{tab:inference_speed}
\end{table}

\subsection{Compositionality of Viability Filters}
\label{sec:inference_speed}
We can compose $\VF$s to tackle settings where the character encounters multiple tasks in the same environment as shown in Fig.~\ref{fig:inference}. The single diffusion planner provides a plan, and the resulting~$\VF$ is the multiplication of the task-specific~$\VF$ as given by:
\begin{equation}
    \VF(p_t) = \VF_{p}(s_{t,p},p_t)\times \VF_h(s_{t,h},p_t) \times \VF_o (s_{t,o},p_t)
\label{eq:composition_vf}
\end{equation}

where $p_t$ denotes the generated plan, $\VF_p,\VF_h,\VF_o$ and $s_{t,p},s_{t,h},s_{t,o}$ correspond to the $\VF$ and the state for the platform, hurdle, and obstacle tasks respectively. A video demonstration of the character navigating in this setting can be found in our supplementary materials.

\section{Conclusions}    

We have proposed a diffusion-based planner enhanced via learned viability filters, trained in an online setting for maximum performance.
Operating together, these enable improved generation of plans that meet implicit feasibility constraints as well as modeling delayed-but-inevitable infeasibility or constraint violation. The viability filter captures these properties in expectation, and can therefore better cope with stochastic environments as compared to a diffusion planner alone.
The viability filters can be asymmetric with respect to the diffusion planner, i.e., they can be conditioned on different observations,
and they can further be effectively composed to satisfy multiple constraints. 
The approach is shown to be faster to compute than guidance-based diffusion, while also avoiding the challenges of tuning guidance-based approaches.
We demonstrate the results on challenging footstep-planning tasks for humanoid movement.


The method builds on elements of diffusion-based control, value-based RL, sampling-based MPC, and asymmetric actor-critic policies.
Exploration is accomplished via already-plausible plans coming from the diffusion model,
while enabling further improvements. 
We expect that similar may prove to be useful for modeling more general reward functions, i.e., considering more than 
simple viability. 
In the context of RL, a limitation of the current work is that no explicit exploration is carried out 
beyond what is ``in distribution'' for the diffusion model. An alternative approach may be to allow for some explicit
exploration and then being able to reincorporate positive outcomes back into the diffusion model, in an iterative process.

\clearpage
\bibliography{refs}
\bibliographystyle{arxiv}

\newpage
\appendix
\onecolumn
\clearpage

\section{Network Architecture}
\label{sup:network_architecture}
\subsection{Diffusion Planner}
For the diffusion planner we use a 1D U-Net as described in~\citep{janner2022diffuser} with the $\epsilon$-prediction variant. For conditioning the diffusion model, although many works employ classifier-free sampling~\citep{ho2022classifier} by incorporating both a conditional and unconditional model, we found that the conditional model by itself generated diverse enough samples for our tasks, so for simplicity we only used that one. To extract features for the different conditional inputs, we use a separate MLP for the character state and waypoint, and a ResNet18~\citep{he2016deep} for the heightfield. Finally, we concatenate the extracted features from the conditional input and the denoising step $k$ and add it to the intermediate plan features within each convolutional block similar to \citep{rempeluo2023tracepace,janner2022diffuser,ho2022video}. Tab.~\ref{tab:obs_space_diffusion} and Tab.~\ref{tab:params_dp} include complementary details on the diffusion planner input dimensions and hyperparameters.

\begin{table}[!h]
	\caption{Input Shapes to diffusion planner}
	\centering
	\begin{tabular}{ll}
        \toprule
        \textbf{Name} & \textbf{Shape} \\
        \midrule
        plan  & $4\times3$\\
		heightfield  & $32\times32$ \\
        waypoint position  & 3 \\
        \midrule
        \multicolumn{2}{c}{Character State} \\
        \midrule
		roll, pitch, yaw & 3\\
        body velocities & 3 \\
        joint angles \& velocities & 42\\
        feet contacts & 2\\
        feet position & 6\\
		\bottomrule
	\end{tabular}
	\label{tab:obs_space_diffusion}
\end{table}

\begin{table}[!h]
    \centering
    \caption{Hyperparameters for Diffusion Planner}
    \renewcommand{\arraystretch}{1.2}
    \begin{tabular}{l|c}
        \toprule
        \textbf{Parameter} & \textbf{Diffusion Planner} \\
        \midrule
        Batch size & 256 \\
        Samples trained & 750k \\
        $\beta$ scheduler & Cosine \citep{nichol2021improved} \\
        Learning rate & 2e-5 \\
        Optimizer & Adam \citep{kingma2014adam} \\
        diffusion steps & 20 \\
        Diffusion loss & $\epsilon$ prediction \\
        Diffusion var. & Fixed small $\tilde{\beta_t} = \frac{1 - \alpha_{t-1}}{1 - \alpha_t} \beta_t$ \\
        \bottomrule
    \end{tabular}
    \label{tab:params_dp}
\end{table}

\subsection{Viability Filter}
For the $\VF$,  each distinct input (i.e., heightfield, obstacle, avoidance zone, waypoints, and plan) is first processed through its own embedding network to extract features. For all but the heightfield, this feature extraction is performed using an MLP, while the heightfield is processed using a ResNet18 network. The extracted features are then concatenated and passed through a final MLP, which outputs the expected return for the given plan and state. Tab.~\ref{tab:obs_space_vf} and Tab.~\ref{tab:params_vf} include complementary details on the input dimensionality across the different viability filters and hyperparameters.

\begin{table}[H]
	\caption{Input Shapes for Viability Filters}
	\centering
	\begin{tabular}{ll}
        \toprule
        \textbf{Name} & \textbf{Shape} \\
        \midrule
        \multicolumn{2}{c}{Character State} \\
        \cmidrule(r){1-2}
		roll, pitch, yaw & 3\\
        body velocities & 3 \\
        joint angles \& velocities & 42\\
        feet contacts & 2\\
        feet position & 6\\
        \midrule
        \multicolumn{2}{c}{\textbf{VF Platform}} \\
        \midrule
		heightfield  & $32\times32$ \\
        \midrule
        \multicolumn{2}{c}{\textbf{VF Hurdle}} \\
        \midrule
        hurdle position & 3 \\
        hurdle angle & 1 \\
        \midrule
        \multicolumn{2}{c}{\textbf{VF Obstacle}} \\
        \midrule
        obstacle position & 3 \\
        obstacle radius & 1 \\
		\bottomrule
	\end{tabular}
	\label{tab:obs_space_vf}
\end{table}

\begin{table}[H]
    \centering
    \caption{Hyperparameters for Viability Filter in Offline and Online setting}
    \renewcommand{\arraystretch}{1.2}
    \begin{tabular}{l|c}
        \toprule
        \textbf{Parameter} & \textbf{Viability Filter Offline} \\
        \midrule
        Batch size & 512 \\
        Samples trained & 2M \\
        Learning rate & 1e-4 \\
        Optimizer & Adam \citep{kingma2014adam} \\
        discount factor $\gamma$ & 0.75 \\
        loss & MSE\\
        \midrule
        \textbf{Parameter} & \textbf{Viability Filter Online} \\
        \midrule
        Batch size & 256 \\
        Episodes  & 100k \\
        Environment Horizon & 1000 \\
        Learning rate & 1e-4 \\
        Optimizer & Adam \citep{kingma2014adam} \\
        discount factor $\gamma$ & 0.75 \\
        Experience Replay Size & 100k \\
        $\tau$ & 0.01 \\
        Diffusion Generated Samples & 200 \\
        \bottomrule
    \end{tabular}
    \label{tab:params_vf}
\end{table}

\section{Training Details}

\subsection{Procedural Generator}
\label{supsub:procedural}
To collect the procedurally generated data, we start with a flat terrain and an initial heading angle $\phi_{t}$. The next target footstep along the trajectory is generated by first sampling a new radial footstep distance $\Delta r\sim U(\Delta r_{min},\Delta r_{max})$ and a heading angle offset $\Delta\phi\sim U(-\Delta\phi_{max},\Delta\phi_{max})$. Then, the relative distance to this new target footstep is:
\[
\begin{split}
& \Delta x = \Delta r \cdot \sin(\Delta\phi - \phi_t)\\
& \Delta y = \Delta r \cdot \cos(\Delta\phi + \phi_t).
\end{split}
\]

we rotate the target step's coordinates depending on whether the left or right leg will be used to reach it.

\[
\begin{bmatrix}
\Delta x' \\ 
\Delta y'
\end{bmatrix}
=
\textit{rot}(l_{t+1}) \cdot
\begin{bmatrix}
\Delta x \\ 
\Delta y
\end{bmatrix}
\]

where $\textit{rot}$ is a fixed rotation matrix and $l_{t+1}$ is a binary index corresponding to the left or right leg. The footstep position in global coordinates and new heading angle then become:
\begin{equation}
\begin{split}
& x_{t+1} = x_{t} + \Delta x'\\
& y_{t+1} = y_{t} + \Delta y'\\
& \phi_{t+1} = \phi_{t} + \Delta \phi.
\end{split}
\end{equation}
This is repeated until the full trajectory is generated totaling to 50 footsteps.

\subsection{Diffusion Planner}
\label{supsub:diffusion_training_details}
To train the diffusion planner we collect 500k data points each including: plan, heightfield, character state, waypoint, and return using trajectories from the procedural generator. The return corresponds to the discounted reward for a plan and is not used for the diffusion planner training. From the 500k data points 150k of them are failed plans and 350k are successful. We store all data but use only the successful plans to train the diffusion model. When collecting the data 100k samples come from purely flat terrain, and the remaining 400k come from the platform task. When sampling heights $h$ for the platforms we collect four 100k batches of data where in each batch we reduce the lower bound for the height distribution of the platforms i.e. batch 1: $h\sim U(10,65)$cm, batch 2: $h\sim U(30,65)$cm, batch 3: $h\sim U(40,65)$cm, and batch 4L $h\sim U(50,65)$cm. This is because we want to oversample successful examples where the character traverses the more challenging heights. For the step generation parameters of the procedural generator as referenced in Sup.~\ref{supsub:procedural}, for the radial footstep distance we use use $\Delta r_{min} = 0.5m$ and $\Delta r_{max} = 1.15m$, and for the heading angle offset $\Delta \phi_{max} = 20^\circ$. These values were empirically determined and reflect the limits of the current control policy i.e. narrowest, widest step it can do.

\subsection{Viability Filter}
\paragraph{Offline Setting}
For the platform task we use the data that was collected to train the diffusion planner as referenced in~\ref{supsub:diffusion_training_details} but with both the successful and failed plans. For the hurdle task, we collect 100k samples for each height (i.e. 25cm, 30cm, 35cm) and train a \VF. Note that the data collected for the hurdle task is not used in the diffusion model. 
\vspace{-7pt}
\paragraph{Online Setting}
We describe our training procedure for the online setting on Alg.~\ref{alg:online training} which remains the same across all tasks. For the platform task we choose to bootstrap the \VF using the weights from the offline \VF. Although this step is not necessary it helps to significantly speed up training. For the hurdle and obstacle task we train the online \VF from scratch. We train across all tasks for 100k episodes and update the weights at the end of each episode. For the Experience Replay we use a buffer $D_s$ to store successful transitions i.e. transitions with reward of 1 and a buffer $D_f$ to store failed transitions. During the $\VF$ update half of the sampled batch comes from the success buffer and half from the failure buffer. We do this because the reward signal is sparse (i.e. you fail once within an episode). 

\begin{algorithm}[!h]
   \caption{Viability Filter Online Training}
   \label{alg:online training}
\begin{algorithmic}
   \STATE Initialize success and failure buffer $D_s,D_f$ to capacity $N$, soft-update parameter $\tau$
   \STATE Initialize Viability Filter $\VF_{\theta}$ , target Viability Filter $\hat{\VF}_{\phi}$ and load Diffusion Planner $G_p$
   \STATE Initialize Humanoid Environment
   \FOR{$N_{E}$ episodes}
       \STATE Get planner initial conditioning signal $\rvc_0$ and $\VF$ state $s_{0}$ from environment
       \STATE Generate $N_s$ plans $P_0 \sim G_p(\rvc)$
   \FOR{each timestep $t$ in episode}
       \STATE With probability $\epsilon$, select a random plan $p_t$ where $p_t \in P_t$
       \STATE Otherwise, select $p_t = \arg\max_{p} VF_{\theta}(s_t, P_t)$
       \STATE Execute first target step from $p_t$ in the environment's control loop
       \IF{target step reached without early termination}
           \STATE assign success reward $r_s = 1$, and get $\rvc_{t+1}, s_{t+1}$
           \STATE sample new plans $P_{t+1} = G_p(\rvc_{t+1})$
           \STATE Store transition $(s_t, p_t, r_t, s_{t+1}, P_{t+1})$ in $D_s$
       \ELSIF{entered early termination}
           \STATE assign failure reward $r_f = 0$
           \STATE Store transition $(s_t, p_t, r_f, \emptyset, \emptyset)$ in $D_f$
           \STATE Terminate episode
       \ENDIF
   \ENDFOR
   \STATE Sample a random batch of transitions $(s_i, p_i, r_i, s_{i+1}, P_{i+1})$ equally from $D_s$ and $D_f$
   \STATE Set target $y_i = 
       \begin{cases} 
           r_i & \text{if $s_{i+1}$ is $\emptyset$} \\
           r_i + \gamma \max_{P_{i+1}} \hat{VF}_{\phi}(s_{i+1}, P_{i+1}) & \text{otherwise}
       \end{cases}$
   \STATE Perform a gradient descent step on $(y_i - VF_\theta(s_i, p_i))^2$ with respect to $\theta$
   \STATE Perform soft-update to target network $\phi \gets (1-\tau)\cdot\phi + \tau\cdot\theta$
   \ENDFOR
\end{algorithmic}
\end{algorithm}

\subsubsection{Humanoid Controller}
We train a reinforcement learning (RL) control policy for stepping stone skills, following the work of ALLSTEPS~\cite{2020-SCA-ALLSTEPS}. Although the stepping stone environment differs significantly from our tasks, the policy, relying solely on proprioceptive information, generalizes well to flat terrain and platforms. We utilize PPO~\cite{schulman2017proximal} as our RL training algorithm. Additionally, we adopt the uniform curriculum for stone generation and use the same network architecture described in ALLSTEPS~\cite{2020-SCA-ALLSTEPS}.

The humanoid character's action space is 21-dimensional, comprising joint torques. Its observation space includes joint angles and velocities relative to the parent link, the roll and pitch of the root orientation in global space, linear velocities of the body in the character's root frame, the pelvis height relative to the current stance foot, and binary contact indicators for each foot.

\section{Success Rates}
We also include the results from Fig.~\ref{fig:box success rates} in table form shown in Tab.~\ref{tab:merged_success_rates}.

\begin{table*}[!h]
\caption{Mean success rate and standard deviation per platform from the data in Fig.\ref{fig:box success rates}}
\centering
\begin{tabular}{c|c c c c}
\toprule
\textbf{Platform} & \textbf{Procedural} & \textbf{Diffusion}& \textbf{Diffusion} & \textbf{Diffusion} \\
\textbf{Heights (cm)} & \textbf{Generator} & & \textbf{VF-Offline} & \textbf{VF-Online}\\
\hline
0.10 & 0.88 $\pm$ 0.06 & 0.97 $\pm$ 0.04 & 0.98 $\pm$ 0.02 & 0.97 $\pm$ 0.02 \\
0.15 & 0.90 $\pm$ 0.07 & 0.96 $\pm$ 0.02 & 0.99 $\pm$ 0.02 & 0.99 $\pm$ 0.02 \\
0.20 & 0.91 $\pm$ 0.02 & 0.93 $\pm$ 0.05 & 0.98 $\pm$ 0.02 & 0.98 $\pm$ 0.04 \\
0.25 & 0.81 $\pm$ 0.11 & 0.93 $\pm$ 0.07 & 0.95 $\pm$ 0.08 & 0.96 $\pm$ 0.04 \\
0.30 & 0.67 $\pm$ 0.08 & 0.83 $\pm$ 0.09 & 0.86 $\pm$ 0.06 & 0.95 $\pm$ 0.03 \\
0.35 & 0.55 $\pm$ 0.13 & 0.80 $\pm$ 0.08 & 0.95 $\pm$ 0.04 & 0.93 $\pm$ 0.04 \\
0.40 & 0.62 $\pm$ 0.07 & 0.87 $\pm$ 0.09 & 0.94 $\pm$ 0.02 & 0.95 $\pm$ 0.03 \\
0.45 & 0.41 $\pm$ 0.10 & 0.79 $\pm$ 0.06 & 0.95 $\pm$ 0.05 & 0.96 $\pm$ 0.04 \\
0.50 & 0.36 $\pm$ 0.10 & 0.81 $\pm$ 0.09 & 0.97 $\pm$ 0.04 & 0.97 $\pm$ 0.04 \\
0.55 & 0.33 $\pm$ 0.05 & 0.76 $\pm$ 0.07 & 0.93 $\pm$ 0.04 & 0.98 $\pm$ 0.04 \\
0.60 & 0.18 $\pm$ 0.02 & 0.64 $\pm$ 0.05 & 0.82 $\pm$ 0.07 & 0.88 $\pm$ 0.04 \\
0.65 & 0.05 $\pm$ 0.03 & 0.21 $\pm$ 0.05 & 0.39 $\pm$ 0.06 & 0.68 $\pm$ 0.07 \\
\bottomrule
\end{tabular}
\label{tab:merged_success_rates}
\end{table*}

\newpage

\section{Diffusion Guidance}
\label{sec:appendix-guidance}
\subsection{Classifier-free Guidance}
To provide flexibility over the relative strength of the conditioning signal when using explicit conditional training, \emph{classifier-free guidance} jointly trains an unconditional model $\rveps_\theta(\rvtau^i, i)$ alongside $\rveps_\theta(\rvtau^i, i, \rvc)$ by setting $\rvc=\emptyset$ for a fraction of training samples, e.g. $10\%$. At inference time, a weighted combination of the two models is used to trade-off between fidelity and diversity using $w$:
    \begin{align*}
        \rveps_\theta(\rvtau^i, i, \rvc) = {\rveps_\theta(\rvtau^i, i, \emptyset)} + w \left( {\rveps_\theta(\rvtau^i, i, \rvc)} - {\rveps_\theta(\rvtau^i, i, \emptyset)} \right).
    \end{align*}

\subsection{Classifier Guidance}
Classifier guidance allows conditioning an unconditionally trained diffusion model at inference time. Having a differentiable function $f_\phi(\rvtau^i)$ that provides meaningful feedback about the denoised sample (e.g. loss, or reward functions, or a classifier), classifier guidance pushes the mean estimate of the reverse process towards the gradients of $f$. Since our diffusion model outputs noise estimates, we need to first reparameterize the output to mean estimates. This is done as below following \citet{ho2020denoising}:
    \begin{align}
        \tilde{\rvmu}(\rvtau^i, i) = \frac{1}{\sqrt{\alpha^i}} \left ( \rvtau^i - \frac{\beta^i}{\sqrt{1-\bar{\alpha}^i}} \rveps_\theta(\rvtau^i, i) \right ).
    \end{align}
Now, without guidance the denoising would be done by simply following:
    \begin{align}
        \rvtau^{i-1} \sim \mathcal{N} (\tilde{\rvmu}(\rvtau^i, i), \rvsigma^i).
    \end{align}
Classifier guided denoising will modify this process as below following~\citet{dhariwal2021diffusion}:
    \begin{align}
        \rvtau^{i-1} \sim \mathcal{N} (\tilde{\rvmu}(\rvtau^i, i) + w\rvsigma^i\nabla_{\rvtau^i}\log f_\phi(\rvtau^i), \rvsigma^i)
    \end{align}
where $w$ is the gradient scale.

\end{document}